\pdfoutput=1
\documentclass[11pt]{article}
\usepackage{acl}
\usepackage{caption,subcaption}
\usepackage{times}
\usepackage{latexsym}
\usepackage[T1]{fontenc}
\usepackage[utf8]{inputenc}
\usepackage{microtype}
\usepackage{hyperref}
\usepackage{amsmath}
\usepackage{amssymb}
\usepackage{natbib}
\usepackage{bm}
\usepackage{cleveref}
\usepackage{upgreek}
\usepackage{mathrsfs}
\usepackage{mathtools}
\usepackage{nccmath}
\usepackage{url}
\usepackage{amssymb}
\usepackage{amsthm}
\newtheorem{definition}{Def}
\usepackage[inline]{enumitem}
\usepackage{makecell}
\usepackage{boldline}
\usepackage{booktabs}
\usepackage{arydshln}
\usepackage{multirow}
\usepackage[htt]{hyphenat}
\usepackage{soul}
\usepackage{color, colortbl}
\usepackage{xcolor}		

\usepackage{tikz}
\newcommand*\circled[1]{\tikz[baseline=(char.base)]{
            \node[shape=circle,draw,inner sep=1pt] (char) {#1};}}
\newcommand\aspect{\textsc{Aspire}}
\newcommand{\RNum}[1]{\uppercase\expandafter{\romannumeral #1\relax}}

\usepackage{todonotes}
\newcommand{\Note}[2]{}
\definecolor{brightmaroon}{rgb}{0.76, 0.23, 0.28}

\newcommand{\tom}[1]{\Note{blue!30}{#1 --Tom}}
\newcommand{\smysore}[1]{\Note{purple!30}{#1 --SM}}

\title{Multi-Vector Models with Textual Guidance for \\  Fine-Grained Scientific Document Similarity}

\author{
    Sheshera Mysore\textsuperscript{1}\thanks{* Part of the work done during internship at AI2.}\quad
    Arman Cohan\textsuperscript{2,3} \quad
    Tom Hope\textsuperscript{2,3}\\
  \textsuperscript{1}University of Massachusetts Amherst, \textsc{\small MA, USA} \quad \textsuperscript{2}University of Washington, \textsc{\small WA, USA}\\ \textsuperscript{3}Allen Institute for Artificial Intelligence, \textsc{\small WA, USA}\\
  \texttt{smysore@cs.umass.edu}\quad \texttt{\{tomh,armanc\}@allenai.org} \\ }

\begin{document}
\maketitle

\begin{abstract}
We present a new scientific document similarity model based on matching fine-grained aspects of texts. To train our model, we exploit a naturally-occurring source of supervision: sentences in the full-text of papers that cite multiple papers together (\emph{co-citations}). Such co-citations not only reflect close paper relatedness, but also provide textual descriptions of \emph{how} the co-cited papers are related. This novel form of \emph{textual supervision} is used for learning to match aspects across papers. We develop multi-vector representations where vectors correspond to sentence-level aspects of documents, and present two methods for aspect matching: (1) A fast method that only matches single aspects, and (2) a method that makes sparse multiple matches with an Optimal Transport mechanism that computes an Earth Mover's Distance between aspects.
Our approach improves performance on document similarity tasks in four datasets. Further, our fast single-match method achieves competitive results, paving the way for applying fine-grained similarity to large scientific corpora.\footnote{Code, data, and models available at: \url{https://github.com/allenai/aspire}}

\end{abstract}

\section{Introduction}
\label{sec-intro}
The ability to identify similarity across documents in large scientific corpora is fundamental for many applications, including recommendation \cite{bhagavatula2018content}, exploratory or analogical search \cite{hope2017accelerating,hope2021scaling, lissandrini2019exampletutorial}, paper-reviewer matching 
\cite{mimno2007expertise, berger2020effective} and many more uses. 

Scientific papers often describe multifaceted arguments and ideas \cite{hope-etal-2021-extracting,lahav2021search}, suggesting that models capable of matching specific aspects can better capture overall document relatedness, too. For example, sentences in research abstracts can often be categorized as descriptions of objectives, methods, or findings \cite{kim2011automatic, chan2018solvent}, centrally important discourse structures of scientific texts.

In this paper, we propose a new model for document similarity that makes aspect-level matches across papers and aggregates them into a document-level similarity. We focus on sentence-level aspects of paper abstracts, and train multi-vector representations of papers in terms of their contextualized sentence embeddings. To train our models, we leverage a readily available data source: sentences that co-cite multiple papers. Unlike recent work that used citation links for learning scientific document similarity \cite{cohan2020specter}, we observe that papers cited in close proximity provide a more precise indication of relatedness. Furthermore, the citing sentences typically describe how the co-cited papers are related, in terms of shared aspects (e.g., similar methods or findings, related challenges or directions, etc.). Building on this observation, we leverage these textual descriptions as a novel source of \emph{textual supervision}, using them to guide our model to learn which sentence-aspects match without any direct sentence-level supervision. Guidance for the document similarity model is obtained via an auxiliary sentence encoder model that is used for aligning abstract sentences by finding pairs most similar to the citing sentence text.

Our document similarity objective is modeled as a function of similarity between sentence-level matches. We explore two strategies to aggregate over sentence-level distances between documents. First, a single-match method with minimum L2 distances between document aspect vectors. This approach readily supports approximate nearest neighbor search methods for large-scale retrieval. Second, a multi-match method that computes an Earth Mover's Distance between documents' aspect vectors by solving an Optimal Transport problem. This yields a soft sparse matching of aspect vectors, which when combined with their L2 distances gives a document-level distance.

Finally, as an additional benefit of our representation, our models also support a finer \textit{aspect-conditional} retrieval task \cite{hope2017accelerating,hope-etal-2021-extracting,chan2018solvent,MysoreCSFCUBE2021} where aspects can be specified by selecting abstract sentences --- for example, selecting sentences describing methods and retrieving papers using similar methods. As we show, naively encoding sentences without their context leads to subpar results in this task, and our representation that does take context into account dramatically improves results.

Extensive empirical evaluation on four English scientific text datasets and seven similarity tasks at the level of documents and sentences demonstrates the effectiveness of our models. These include biomedical document retrieval tasks and a recent faceted query-by-example corpus of computer science papers \cite{MysoreCSFCUBE2021}. This latter dataset is used for evaluating retrieval conditioned on specific aspects in context (e.g., for finding papers with similar methods to a query document), demonstrating that our model can be used in this challenging and important setting. In summary, we make the following main contributions: 

\begin{enumerate}[nosep]
    \item \textbf{Multi-Vector Document Similarity Model}: 
    We present \aspect\footnote{\aspect: \ul{Asp}ectual Sc\ul{i}entific Paper \ul{Re}lations.}, a multi-vector document similarity model that flexibly aggregates over fine-grained sentence-level aspect matches.
    \item \textbf{Co-Citation Context Supervision}:
    We exploit widely-available co-citation sentences as a new source of training data for document similarity and provide a method using a novel form of textual supervision to guide representation learning for aspect matching. 
    \item \textbf{State of the Art Results}: Our \aspect~models outperforms strong baseline methods across four datasets for the abstract  and aspect-conditional similarity tasks.
\end{enumerate}

\begin{figure*}[t]
     \centering
     \includegraphics[width=0.8\textwidth]{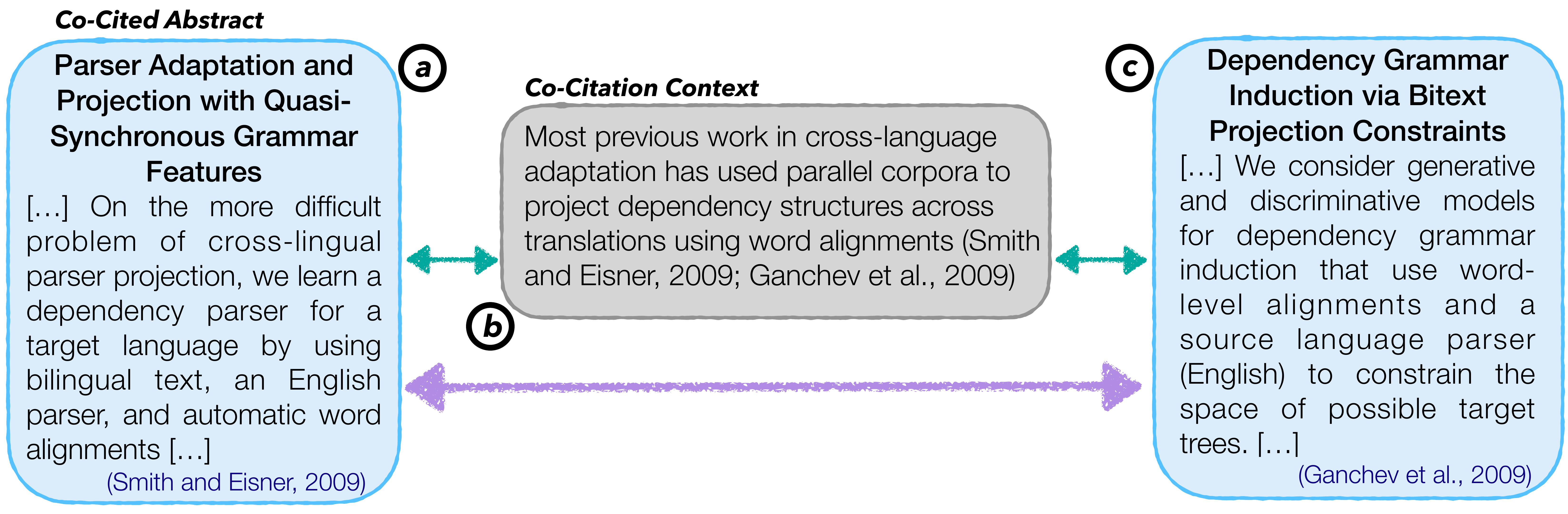}
     \caption{Example illustrating the signal in co-citations. Of all the sentences in co-cited abstracts \circled{\small a} and \circled{\small c}, the sentences shown are each individually aligned to co-citation context \circled{\small b} as per embeddings from $\mathtt{BERT}_{\mathcal{E}}$ (\S\ref{sec-single-match}). Consequently these sentences in \circled{\small a} and \circled{\small c} are treated as sharing aspects between the co-cited papers and our fine-grained similarity model for single matching is trained to align them.}
     \label{fig-text-example}
 \end{figure*}
\section{Problem Setup}
\label{sec-problem-setup}\tom{I don't think we need all this (plus it's too plagiarized from CSFcube :-). Just define Q and C in the model. the stuff about ranking is never really used by us in the paper - right?. it may raise q's on the disconnect between this and what we do. }
\smysore{Removed a bunch of stuff copied from csfcube and the ranking stuff, not sure i follow what the disconnect is tho.} 
Given query document $Q$ and a candidate amongst a set of documents $C \in \mathcal{C}$, where documents consists of $N$ sentences $\langle S_1, S_2, \dots S_N\rangle$ we aim to leverage fine-grained document similarity in two problem settings. An abstract level retrieval task \cite{Brown2019RELISH,cohan2020specter} and an aspect-level retrieval task \cite{MysoreCSFCUBE2021}:

\begin{definition} Retrieval by abstracts:
Given query and candidate documents -- $Q$ and $\mathcal{C}$ a system must
output the ranking over $\mathcal{C}$.
\label{def-query-by-abstract}
\end{definition}

\begin{definition} 
Aspect-level retrieval by sentences:
Given query and candidate documents -- $Q$ and $\mathcal{C}$, and a subset of
sentences $\mathcal{S}_Q\subseteq Q$ conditional on which to retrieve documents, a system must output the ranking over $\mathcal{C}$.
\label{def-query-by-sentence}
\end{definition}


\textbf{Modeling Desiderata:}
Next, we also outline key desired properties we require from models developed for task definitions \ref{def-query-by-abstract} and \ref{def-query-by-sentence}. We follow these desiderata when building our methods (\S\ref{sec-model-general}).

\begin{enumerate*}[nolistsep, noitemsep]
    \item \textit{Allowing specification of optional fine-grained aspects}: We would like models to allow the ability to specify fine-grained query aspects in a query document based on which retrievals should be made. These may be obtained automatically (e.g., with a discourse tagging method) or via user specification.\\
    \item \textit{Scalable to large corpora and efficient inference}:
    State of the art retrieval systems often rely on expensive cross-attention mechanisms on query-document pairs making training and inference expensive \cite{zamani2018noreranking, lin2021pretrained}. This is exacerbated for longer scientific documents requiring specific transformer models \cite{caciularu2021cdlm}. We require our methods to leverage large training corpora and allow efficient inference at scale.
\end{enumerate*}
\section{Proposed Approach: \aspect}
In this section we describe our approach to document similarity -- \aspect. We model finer-grained matches between documents at the level of sentences via contextual representations and aggregating over matches to obtain similarities between whole documents.
We leverage co-citation sentences as a source of document similarity and also as implicit \emph{textual supervision} describing related aspects of co-cited documents. We formulate our multi-vector models \cite{luan2021sparse,Humeau2020Polyencoders} that can support scalable inference as novel multiple-instance learning (MIL) models.

\subsection{Fine-grained Document Similarity}
\label{sec-model-general}
We assume to be given a training set consisting of sets of documents $\mathcal{P}$ which are \emph{weakly-labeled} for similarity. We leverage widely-available sets of papers co-cited together in the same sentence as similar (see Figure \ref{fig-text-example}). This builds on the observation that co-citations in close proximity (e.g., in the same sentence) are strong indicators for paper relatedness \cite{gipp2009citation}. 

We follow the contrastive learning framework, commonly used for learning semantic similarity \cite{ reimers2019sentencebert,cohan2020specter}. We train models on triples of the form $(p, p', n)$ where $p,p'\in \mathcal{P}$ and $n\notin\mathcal{P}$ is a randomly selected negative, using the triple margin ranking loss $\mathcal{L}_{f}(p,p',n) = \mathtt{max}[f(p, p')-f(p,n)+m,0]$, where $f(\cdot,\cdot)$ is a distance between documents. 
All pairwise-combinations $p,p'\in \mathcal{P}$ are treated as positive pairs in-turn. In this work, we parameterize $f$ based on the distances between finer-grained document aspects $\mathcal{A}$. Given documents $p$ and $p'$, we focus on a family of functions $f$ of the form:
\begin{align}
    f(p,p') = \underset{(i,i')\in\mathcal{A}_p\times\mathcal{A}_{p'}}{\sum} w_{i,i'}\cdot d_{i,i'}.
    \label{eq-general-alignments}
\end{align}
Here, $\mathcal{A}_p\times\mathcal{A}_p'$ represents the space of 
alignments between aspects of document $p$ and $p'$,  $d_{i,i'}$ 
denotes a distance between two aspects $i,i'$, and $w_{i,i'}$ 
represents a weight indicating the contribution of the aspect 
similarity to the overall document similarity. Unlike previous work \cite{neves2019evaluation, jain2018learning, hope2017accelerating}, we make no assumption on specific aspect semantics in deriving a model architecture, and focus on aspects in the form of \emph{general} subsets of document sentences.


For learning, we only assume to be given \emph{document-level} supervision (sets of documents $\mathcal{P}$), and no gold supervision on \emph{aspect-level} similarity as in other related work, eg.\ \citet{jain2018learning}. Our task thus consists of learning $w_{i,i'}$ and $d_{i,i'}$ via indirect supervision. We cast this problem setting as a novel type of multi-instance learning (MIL) \cite{ilse2018attmi} problem. Prior work in MIL broadly aims to learn instance level classifiers given labels for a bag of instances, this bears resemblance to our setting, where instances are aspects $\mathcal{A}$. However, unlike prior MIL work we focus on learning \emph{similarity} rather than  \emph{classification}. We formulate two variants of $f$ in Equation \ref{eq-general-alignments}:

\begin{enumerate*}[noitemsep, nolistsep, label={(\theenumi)}]
    \item A single match model (\S\ref{sec-single-match}) which considers documents similar based on the single most similar alignment $\hat i_p, \hat i_{p^{'}} \in\mathcal{A}_p\times\mathcal{A}_p'$. This  assumes $w=1$ for the best alignment and $w=0$ elsewhere.
\end{enumerate*}

\begin{enumerate*}[noitemsep, nolistsep, label={(\theenumi)}]
    \setcounter{enumi}{1}
    \item A multi match model (\S\ref{sec-multi-align-ot}) which makes multiple alignments between documents. We find aspect importance weights $w_{i,i'}$, by solving an Optimal Transport (OT) problem \cite{peyre2019computational}.
\end{enumerate*}

In both variants, during training we learn contextualized aspect embeddings that minimize the contrastive loss paramertized with $f$, described further in \S\ref{sec-model-descr}.

\begin{figure*}
\centering
\begin{subfigure}[c]{.45\textwidth}
    \includegraphics[width=\textwidth]{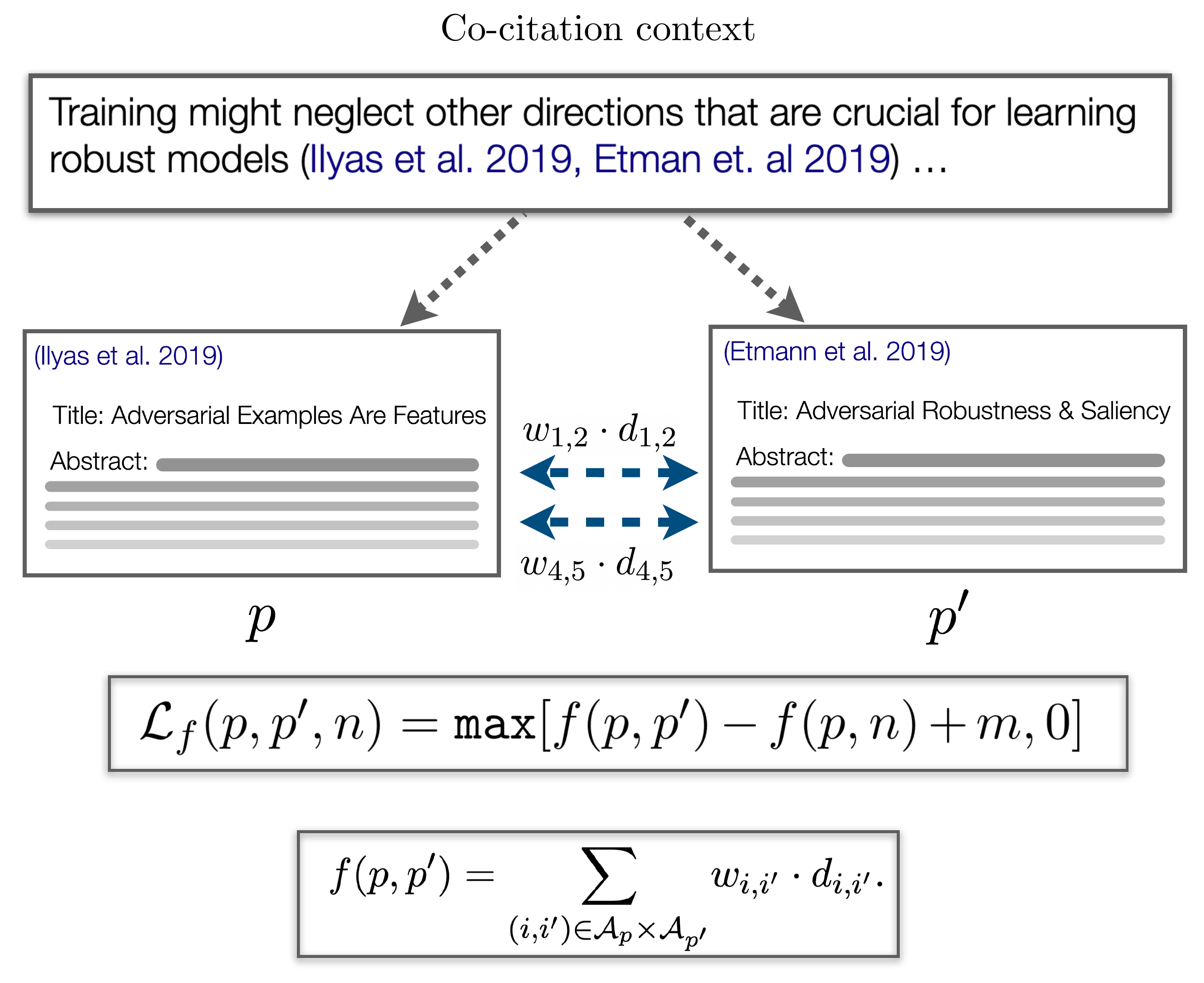}
    \caption{Learning fine-grained document similarity using co-citations.}
    \label{fig-miframework}
\end{subfigure}\quad
\begin{subfigure}[c]{.4\textwidth}
    \includegraphics[width=\textwidth]{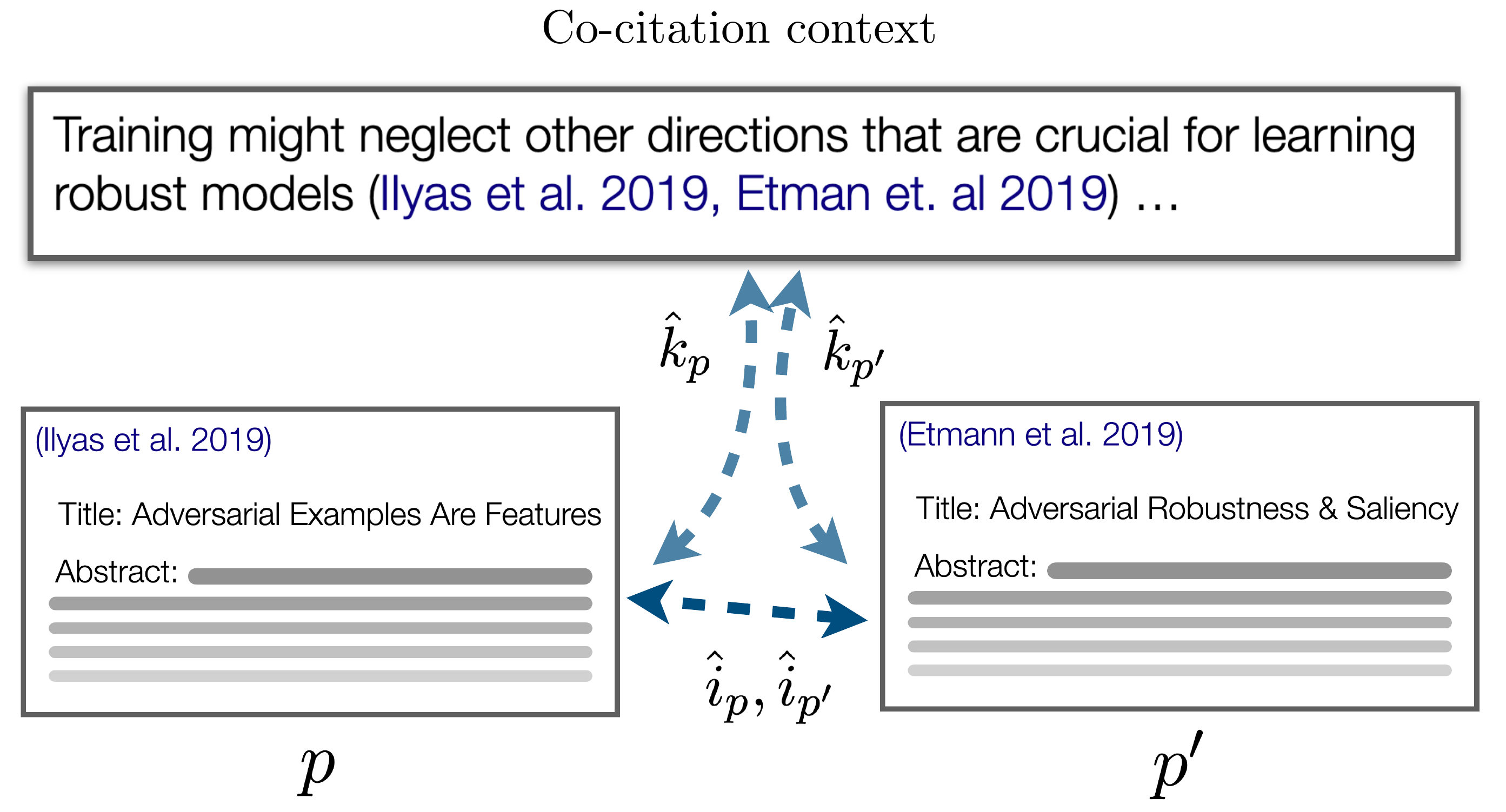}
    \caption{The single best match, $\hat{i}_p,\hat{i}_{p'}$, is computed from textual supervision in the co-citation context.}
    \label{fig-expl-alignment}
    \vspace{2ex}
    \includegraphics[width=\textwidth]{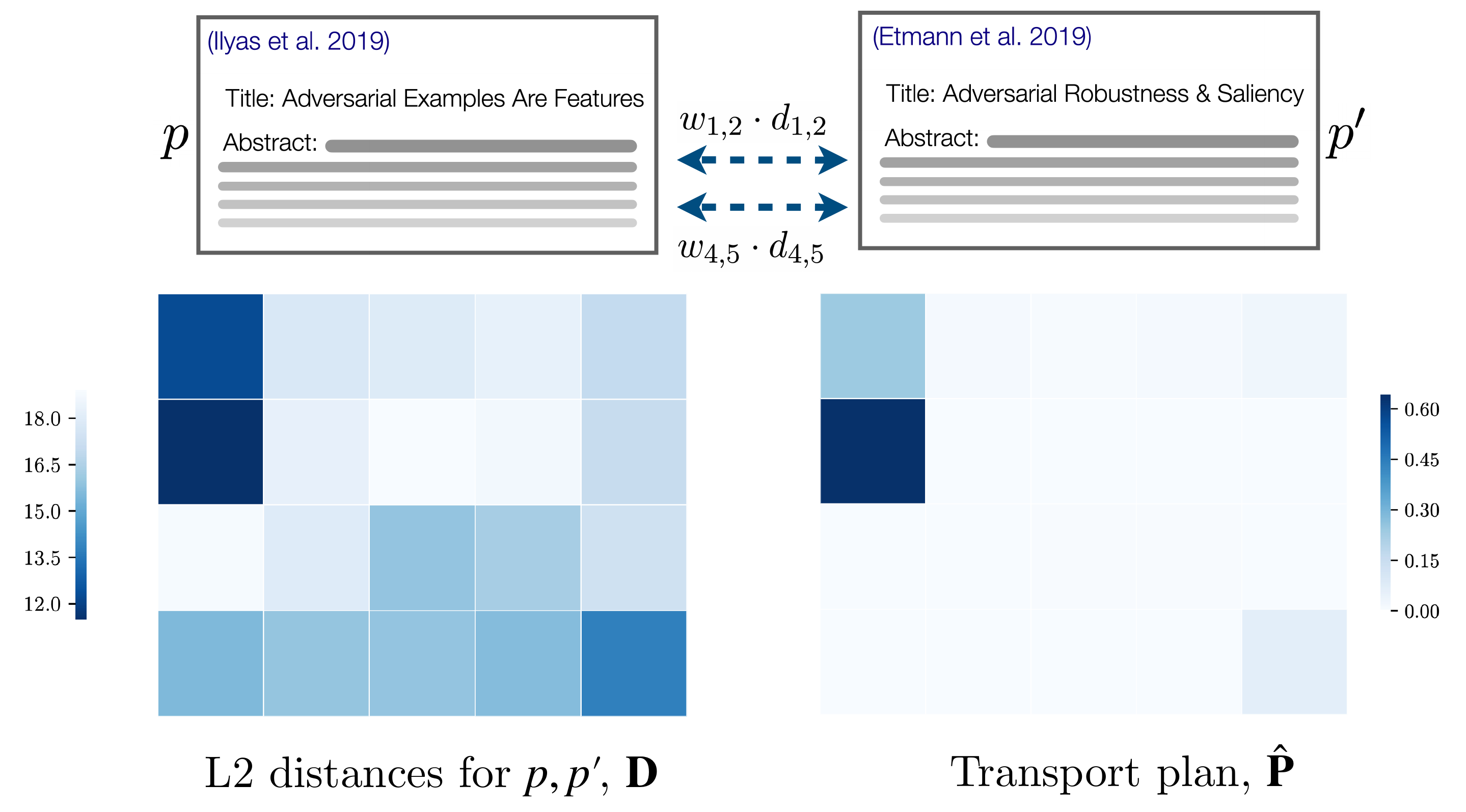}
    \caption{Multi-aspect matches via a sparse transport plan.}
    \label{fig-ot-alignment}
\end{subfigure}
\caption{Approach overview. (a) We train fine-grained similarity models using papers co-cited in the the same sentence in research papers. (b) Single-match models are learned from implicit supervision in co-citation contexts. (c) Multi-match models are learned by aligning aspect representations by solving an Optimal Transport problem.}
\end{figure*}

\textbf{Co-citation Contexts as Supervision:}
\label{sec-fine-grained-data}
Finally, we present a method for incorporating implicit natural language supervision during training, presented by co-citation sentences which describe specific relations between co-cited documents. For example, Figure \ref{fig-text-example} shows a case explaining the similarity between the co-cited papers' methods. We leverage this textual supervision to find a ``best'' alignment $\hat i_p, \hat i_{p^{'}}$ in the single-alignment variant (1), and for guiding the optimal transport plan in variant (2). We describe the specific model components next. Fig \ref{fig-miframework} presents a schematic for our approach.

 
\subsection{Model Description}
\label{sec-model-descr}
\subsubsection{Document Encoder}
\label{sec-doc-encoder}
We leverage a pre-trained BERT-based language model as a document encoder as the base of all our methods. Our encoder is mainly intended to output contextualized 
sentence representations. Given a document title and abstract, this is achieved as:
\begin{align}
    \mathbf{S} = \mathtt{BERT_{\theta}([CLS]~Title~[SEP]~Abstract)}
\label{eq-encoder-params}
\end{align}
where $\mathbf{S} \in \mathbb{R}^{N\times d}$ represents contextualized sentences $\mathbf{s}_1\dots\mathbf{s}_N$
stacked into a matrix.
Here, each $\mathbf{s}$ is
obtained by mean-pooling word-piece embeddings from the final layer of $\mathtt{BERT}_{\theta}$ for the sentence tokens. Pairwise distances between sentences $d_{i,i'}$ in Eq \ref{eq-general-alignments} for $p,p'$, are represented as a matrix $\mathbf{D} \in \mathbb{R}^{N\times N'}$ of L2 distances between $\mathbf{S}_{p}$ and $\mathbf{S}_{p'}$.

\subsubsection{Single Match \& Textual Supervision}
\label{sec-single-match}
Our single match model makes the assumption that document similarity is explained by a single best match, giving $f_{\mathtt{TS}}(p,p')=\mathbf{D}[\hat i_p, \hat i_{p^{'}}]$. Here, we leverage weak supervision from co-citation contexts for training.
This is done by using an auxiliary sentence encoder to compute a maximally aligned sentence $\hat i_p$ in co-cited paper $p$ to the co-citation context, similarly $\hat i_{p'}$ aligns a sentence in $p'$ to the co-citation context. Then the two context aligned sentences are treated as aligned to each other, for training.
In practice, the same papers $\mathcal{P}$ can be co-cited in multiple different papers (in $\sim30\%$ of co-cited papers) giving us a set of co-citation sentences, $e\in\mathcal{E}$ and training data of the form $(\mathcal{E}, \mathcal{P})$.
Alignments of the sentences in 
$p$ and $p'$ to the co-citation contexts $e\in\mathcal{E}$ are computed as:
\begin{align}
\begin{split}
    \hat i_p, \hat k_p &= \underset{i=1\dots N, k=1\dots N'}{\mathtt{argmax}}\mathbf{R}_{p}\mathbf{R}^T_{\mathcal{E}}\\
    \hat i_{p^{'}}, \hat k_{p^{'}} &= \underset{i=1\dots N, k=1\dots N'}{\mathtt{argmax}}\mathbf{R}_{p^{'}}\mathbf{R}^T_{\mathcal{E}}
\end{split}
\label{eq-sb-align}
\end{align}
Here $\mathbf{R}_p$, $\mathbf{R}_{p^{'}}$, and $\mathbf{R}^T_{\mathcal{E}}$ are independent sentence representations for $p, p^{'}$ and $e$, respectively, obtained from a auxiliary sentence encoder $\mathtt{BERT}_{\mathcal{E}}$ (details below), and $\hat{i_p}, \hat{i}_{p^{'}}$ represent the single best alignment of sentences across $p,p'$ ``anchored'' on textual supervision sentences $\mathcal{E}$. Importantly, this supervision is only used during training time to guide learning. This procedure is depicted in Figure \ref{fig-expl-alignment} with Figure \ref{fig-text-example} showing an example. 


\textbf{Co-citation Context Encoder}
\label{subsub:contextenc}
The encoder $\mathtt{BERT}_{\mathcal{E}}$ represents a \textsc{SciBERT} based sentence encoder pre-trained for scientific text similarity. We train  $\mathtt{BERT}_{\mathcal{E}}$ on sets of co-citation contexts referencing 
the same set of papers (i.e. $\mathcal{E}$) in a contrastive learning setup with random in-batch negative samples. This set, $\mathcal{E}$, can be considered as paraphrases since co-citation sentences citing the same papers often describe similar relations between the papers.
This model is similar to Sentence-BERT \cite{reimers2019sentencebert} and we refer to it as 
\texttt{CoSentBert}. In training document encoder $\mathtt{BERT}_{\theta}$, we keep $\mathtt{BERT}_{\mathcal{E}}$ frozen. Appendix \ref{appendix-cosentbert} presents more detail on $\mathtt{BERT}_{\mathcal{E}}$ design. 

\subsubsection{Multiple Matches \& Optimal Transport}
\label{sec-multi-align-ot}
While a single best sentence alignment $\hat{i_p}, \hat{i}_{p^{'}}$ may sufficiently explain document similarity for some documents and applications, documents often have a stronger and weaker alignments. So, in computing sentence alignments between documents we would like a sparse matching that aptly weights alignments while ignoring non-alignments  --- corresponding to learning weights $w_{i,i'}$ in Eq \ref{eq-general-alignments}. To model this intuition we leverage optimal transport.

\textbf{Optimal Transport} The OT problem is constituted by two sets of points, $\mathbf{S}_p$ and 
$\mathbf{S}_{p^{'}}$ as in our case, and distributions $\mathbf{x}_p$ and $\mathbf{x}_{p'}$ according to which the set of points is distributed. The OT problem involves computation of 
a transport plan $\hat{\mathbf{P}}$, which converts $\mathbf{x}_p$ into 
$\mathbf{x}_{p'}$ by transporting probability mass while minimizing an aggregate cost computed from the pairwise costs $\mathbf{D}$ of aligning the points in $\mathbf{S}_p$ and $\mathbf{S}_{p^{'}}$. $\hat{\mathbf{P}}$ is constrained such that its columns and rows 
marginalize respectively to $\mathbf{x}_p$ and $\mathbf{x}_{p'}$ (so that all mass is accounted for).  
Specifically, the computation of $\hat{\mathbf{P}}$ takes the form of a constrained linear optimization problem:
\begin{align}
    {\mathcal{W}} &= \underset{\mathbf{P} \in \mathcal{S}}{\mathtt{min}}\langle\mathbf{D},\mathbf{P}\rangle\\
    &= \underset{\mathbf{P} \in \mathcal{S}}{\mathtt{min}}\overset{N}{\underset{i=1}{\sum}}\overset{N'}{\underset{j=1}{\sum}}\mathbf{D}[i,j]\mathbf{P}[i,j]
    \label{eq-ot-opt}\\
\small
\begin{split}
\mathcal{S}=\{\mathbf{P} \in& \mathbb{R}^{N\times N'}_{+}|\mathbf{P}\mathbf{1}_{N'}
=\mathbf{x}_p, \mathbf{P}^{T}\mathbf{1}_{N'}=\mathbf{x}_{p'}\}    
\end{split}
    \label{eq-plan-constraints}
\end{align} 
%
where $\mathcal{W}$ refers to the Wasserstein or Earth Movers Distance and $\hat{\mathbf{P}}$ is the minimizer resulting from solving Eq \ref{eq-ot-opt}. Of interest here is an established result which shows $\hat{\mathbf{P}}$ to be sparse with $\mathcal{O}(N+N')$ non-zero entries \cite{swanson2020rationalizing}. Therefore, $\hat{\mathbf{P}}$ represents a soft sparse alignment of sentences and can be used as weights $w_{i,i'}$ in Eq \ref{eq-general-alignments}, with document distances computed as $f_{\mathtt{OT}}(p,p') = \langle\mathbf{D},\hat{\mathbf{P}}\rangle$. Fig \ref{fig-ot-alignment} presents a schematic for this approach.

Note that $\mathbf{x}_p$ and $\mathbf{x}_{p'}$ allow control over
importance of sentences in $p$ and $p'$ in the form of relative probability mass. We compute these distributions using pairwise distances as 
$\mathbf{x} = \mathtt{softmax}(-\mathbf{s}/\tau)$ where
$\mathbf{s}_p={\mathtt{min}_i}~\mathbf{D}$ and 
$\mathbf{s}_{p'}={\mathtt{min}_j}~\mathbf{D}$, and $\tau$ is a softmax temperature hyper-parameter. 

For our neural network models trained with automatic differentiation, we leverage an 
entropy regularized version of the Wasserstein distance in Eq \ref{eq-ot-opt} \cite{cuturi2013sinkhorn}. Here 
computation of $\hat{\mathbf{P}}$, is achieved via Sinkhorn iterations, a set of iterative linear updates allowing training with autodiff libraries and leveraging GPU computation. Finally, \citet{cuturi2013sinkhorn} show that computing $\mathcal{W}$ with Sinkhorn iterations shows an empirical quadratic complexity, i.e. $\mathcal{O}(N^2)$ --- similar to that of attention as in a model for late interaction \cite{Humeau2020Polyencoders}.

\textbf{Multi-task model:} To leverage training signals used in both the single and multi-match models, we train a multi match model supervised with textual supervision in a multi-task setup: $\mathcal{L}_{f_{\mathtt{TS}}}+\mathcal{L}_{f_{\mathtt{OT}}}$. 


\subsection{Inference}
\label{sec-inference}
As outlined in \S\ref{sec-problem-setup}, we are interested in a whole-abstract based retrieval (Def \ref{def-query-by-abstract}) and an aspect level retrieval (Def \ref{def-query-by-sentence}). 
In both setups given a query $Q$ and candidate $C$ documents we denote sentence representations from a trained model by $\mathbf{S}_Q$ and $\mathbf{S}_{C}$. For both tasks, we compute distances for ranking while controlling the aspects $\mathcal{A}_Q$ (i.e $\mathcal{A}_p$) over which the weighted sum of Eq \ref{eq-general-alignments} is performed.

\textbf{Whole abstract retrieval:} 
This corresponds to a setup where all aspects of the query document $\mathcal{A}_p$ are used in computing distances between documents. In the single-alignment models, candidates $C$ are ranked based on their maximally aligned sentence with $Q$ using distances from a trained model: $\hat i_p, \hat i_{p^{'}} = \mathtt{argmin}_{i,j}~\mathbf{D}$.
The multi match model ranks candidates using the distance $\langle \mathbf{D}, \hat{\mathbf{P}}\rangle$, where $\hat{\mathbf{P}}$ is the solution to transport problem of Eq \ref{eq-ot-opt}.

\textbf{Aspect level retrieval:} 
In aspect-level retrieval, a subset of sentences $\mathcal{A}_q\subset\mathcal{A}_Q$ is used for query document $Q$; for candidate documents $C$, we do not assume to be given specific aspects, and matching is done across all sentences in each $C$.
In the single alignment models, we only consider a subset of the pairwise sentence distances to determine the 
maximally aligned sentences, giving
$\mathbf{D}^{\mathcal{A}}=\mathbf{D}[\mathcal{A}_q,:]$. This corresponds 
to finding the maximally aligned candidate sentence to the query sentences in  $\mathcal{A}_q$. Similarly, in the
multiple-alignment model we compute the plan $\hat{\mathbf{P}}^{\mathcal{A}_q}$ based on 
the subset of sentences corresponding to $\mathcal{A}_q$ and generate rankings by $
\langle \mathbf{D}^{\mathcal{A}_q}, \hat{\mathbf{P}}^{\mathcal{A}_q}\rangle$. 
Note that $\mathbf{S}_Q$ in $Q$ is still contextualized, capturing document context of sentences not explicitly used in $\mathcal{A}_q$. 

\textbf{Scaling Inference:} Our multi-vector model for single matching performs retrievals via minimum L2 distance. Therefore, this method is amenable to approximate nearest neighbour (ANN) search methods  for large-scale retrieval \cite{andoni2018approximate, luan2021sparse}. Retrieval with our single-match model would involve $|\mathcal{A}_Q|$ and $|\mathcal{A}_q|$ calls to an ANN structure for the whole abstract and aspect-level tasks respectively.

On the other hand, as stated earlier our multi-match model using Sinkhorn iterations involves a $\mathcal{O}(N^2)$ computation \cite{cuturi2013sinkhorn}, which is similar to late interaction methods. \citet{Humeau2020Polyencoders} show late interaction models to be significantly cheaper than cross-encoders while retaining most of their performance in ad-hoc search setups. While quadratic, OT computation in practice can be time-consuming, however, recent work of \citet{backurs2020OTscalable} has seen development of fast ANN methods for Wasserstein distances with practical run-times significantly smaller than quadratic ones. This promises the use of ANN methods in large-scale retrieval with our multi-match model

In our results we refer to our text supervised single match method as \textsc{ts\aspect}, optimal transport multi match method as \textsc{ot\aspect}, and the multi-task trained multi aspect method as \textsc{ts+ot\aspect}.

\section{Experiments and Results}
\label{sec-experiments}
\textbf{Evaluation data:} We evaluate the proposed methods on datasets for whole abstract document similarity and fine-grained document similarity. We overview these below. Appendix \ref{appendix-evaldata} provides detail.

\begin{enumerate*}
    \item \textsc{relish}: An expert annotated dataset of biomedical abstract similarity \cite{Brown2019RELISH}.
    \\
    \item \textsc{treccovid}$_{\mathtt{RF}}$: The original \textsc{treccovid} dataset is labelled for ad-hoc search by experts \cite{treccovid2021}. We reformulate the dataset for abstract similarity, treating all abstracts relevant to one ad-hoc query as similar to each other and dissimilar from abstracts relevant to other queries.
    \\
    \item \textsc{SciDocs}: A benchmark suite of tasks intended for evaluating abstract-level scientific document representations \cite{cohan2020specter}.
    \\
    \item \textsc{CSFCube}: Fine-grained retrieval is evaluated using the recent dataset of \citet{MysoreCSFCUBE2021}, an expert-annotated dataset of machine learning and NLP abstracts labelled against candidates for relevance to  one of 3 broad aspects capturing the main components of methodological research: \texttt{background/objective}, \texttt{method}, \texttt{result}. Relevance is labelled for query sentences corresponding to those aspects, while considering the broader relevance of the sentences' abstract context.
\end{enumerate*}

\begin{table*}[t]
\centering
\scalebox{0.75}{
\begin{tabular}{@{}lcccccccc@{}} \toprule
\textsc{CSFCube} facets $\rightarrow$ & 
\multicolumn{2}{c}{\textit{Aggregated}}   &
\multicolumn{2}{c}{\textit{Background}}       & 
\multicolumn{2}{c}{\textit{Method}}       &
\multicolumn{2}{c}{\textit{Result}}           \\
\cmidrule(lr){2-3} \cmidrule(lr){4-5} \cmidrule(lr){6-7} \cmidrule(lr){8-9}
Models & MAP   & NDCG$_{\%20}$     &  MAP  & NDCG$_{\%20}$ & MAP   & NDCG$_{\%20}$     &  MAP  & NDCG$_{\%20}$  \\\midrule
 \textsc{MPNet-1B}            & 34.64 & 54.94 & 41.06 & 65.86 & 27.21 & 42.48 & 36.07 & 54.94\\
 \textsc{SentBert-PP}            & 26.77 & 48.57 & 35.43 & 60.80 & 16.19 & 33.40 & 29.16 & 48.57\\
 \textsc{SentBert-NLI}           & 25.23 & 45.39 & 30.78 & 54.23 & 15.02 & 31.10 & 30.27 & 45.39\\
 \textsc{UnSimCSE-BERT}          & 24.45 & 42.59 & 30.03 & 51.59 & 14.82 & 31.23 & 28.76 & 42.59\\
 \textsc{SuSimCSE-BERT}          & 23.24 & 43.45 & 30.52 & 55.22 & 13.99 & 30.88 & 25.58 & 43.45\\
 \texttt{CoSentBert}             & 28.95 & 50.68 & 35.78 & 61.27 & 19.27 & 38.77 & 32.15 & 50.68\\
 \textsc{ICTSentBert}            & 28.61 & 48.13 & 35.93 & 59.80 & 15.62 & 35.91 & 34.42 & 48.13\\[2pt]\midrule
 \textsc{otMPNet-1B}             & 36.41 & 56.91 & 43.23 & 67.60 & 28.69 & 43.49 & 37.76 & 60.30\\
 \textsc{Specter}                & 34.23 & 53.28 & 43.95 & 66.70 & 22.44 & 37.41 & 36.79 & 56.67 \\
 \textsc{SciNCL}                & 39.37 & 59.24 & 49.64 & 70.02 & 27.14 & 46.61 & 41.83 & 61.70\\
 \textsc{Specter-CoCite$_\mathtt{Scib}$}   & ${37.90}$ & ${58.16}$ & ${48.40}$ & ${68.71}$ & ${26.95}$ & ${46.79}$ & ${38.93}$ & ${59.68}$\\
 \textsc{Specter-CoCite$_\mathtt{Spec}$}   & ${37.39}$ & ${58.38}$& ${49.99}$ & ${70.03}$ & ${25.60}$ & ${45.99}$ & ${37.33}$ & ${59.95}$\\\midrule
\textsc{ts\aspect}$_\mathtt{Spec}$             & ${40.26}$ & ${60.71}$ & ${49.58}$ & ${70.22}$ & ${\bf 28.86}$ & ${\bf 48.20}$ & ${42.92}$ & ${64.39}$\\
\textsc{ot\aspect}$_\mathtt{Spec}$          & ${\bf 40.79}$ & ${\bf 61.41}$ & ${50.56}$ & ${\bf 71.04}$ & ${27.64}$ & ${46.46}$ & ${\bf 44.75}$ & ${\bf 67.38}$\\
\textsc{ts+ot\aspect}$_\mathtt{Spec}$     & ${40.26}$ & ${60.86}$ & ${\bf 51.79}$ & ${70.99}$ & ${26.68}$ & ${47.60}$ & ${43.06}$ & ${64.82}$
\\\bottomrule
\end{tabular}
}
\caption{Test set results for baseline and proposed methods on \textsc{CSFCube}, an expert annotated fine-grained similarity dataset of computer science papers. Our approaches outperform strong prior models  \textsc{ot}/\textsc{MPNet-1B} and  \textsc{Specter} by 5-6 points, and the concurrently introduced \textsc{SciNCL} model by 1.5-2 points aggregated across queries. Metrics
(MAP, NDCG$_{\%20}$) are computed based on a 2-fold cross-validation and averaged over three re-runs of models. Here, \textsc{ts\aspect}: Text supervised single-match method, \textsc{ot\aspect}: Optimal Transport multi-match method and \textsc{ts+ot\aspect}: Multi-task multi aspect method.}
\label{tab-model-results-compsci}
\end{table*}
\begin{table}[t]
\centering
\scalebox{0.7}{
\begin{tabular}{@{}lcccc@{}}  \toprule
\multirow{2}{*}{Models}& \multicolumn{2}{c}{\textsc{treccovid}$_\mathtt{RF}$}   & \multicolumn{2}{c}{\textsc{relish}}        \\
\cmidrule(lr){2-3} \cmidrule(lr){4-5}
 & MAP   & NDCG$_{\%20}$    &  MAP  & NDCG$_{\%20}$  \\\midrule
 \textsc{MPNet-1B}  & 17.35 & 43.87 & 52.92 & 69.69\\
 \textsc{SentBert-PP}  & 11.12 & 34.85 & 50.80 & 67.35\\
 \textsc{SentBert-NLI} & 13.43 & 40.78 & 47.02 & 63.56\\
 \textsc{UnSimCSE-BERT}  & 9.85  & 34.27 & 45.79 & 62.02\\
 \textsc{SuSimCSE-BERT}  & 11.50 & 37.17 & 47.29 & 63.93\\
\texttt{CoSentBert}     & 12.80 & 38.07 & 50.04 & 66.35\\
 \textsc{ICTSentBert}    & 9.80  & 33.62 & 47.72 & 63.71\\\midrule
 \textsc{otMPNet-1B}     & 27.46 & 58.70 & 57.46 & 74.64\\
 \textsc{Specter}        & 28.24 & 59.28 & 60.62 & 77.20\\
 \textsc{SciNCL}        & 28.73 & 59.16 & 62.09 & 78.72\\
 \textsc{Specter-CoCite$_\mathtt{Scib}$}    & ${30.60}$ & ${62.07}$ & ${61.43}$ & ${78.01}$\\
 \textsc{Specter-CoCite$_\mathtt{Spec}$}    & ${28.59}$ & ${60.07}$ & ${61.43}$ & ${77.96}$\\\midrule
\textsc{ts\aspect}$_\mathtt{Spec}$            & ${26.24}$ & ${56.55}$ & ${61.29}$ & ${77.89}$\\
\textsc{ot\aspect}$_\mathtt{Spec}$         & ${\bf 30.92}$ & ${\bf 62.23}$ & ${62.57}$ & ${78.95}$\\
\textsc{ts+ot\aspect}$_\mathtt{Spec}$    & ${30.90}$ & ${62.18}$ & ${\bf 62.71}$ & ${\bf 79.18}$\\\bottomrule
\end{tabular}
}
\caption{Test set results for baseline and proposed methods on \textsc{treccovid}$_{\mathtt{RF}}$ and \textsc{relish}, expert annotated abstract similarity datasets of biomedical papers. Our approaches outperform or match a strong prior model, \textsc{Specter}, and the concurrently introduced \textsc{SciNCL} by 2-3 points across metrics (MAP, NDCG$_{\%20}$). These are computed as averages over three model re-runs. Method names map similarly to Table \ref{tab-model-results-compsci}.}
\label{tab-model-results-biomed}
\end{table}

\textbf{Baselines:} We compare the proposed approaches to three classes of methods. We overview these classes and associated models below, with Appendix \ref{appendix-baselines} presenting further detail: 
\begin{enumerate*}[noitemsep, nolistsep]
    \item Sentence models: Sentence embedding models present reasonable baselines since we consider fine-grained matches at the sentence level. These are represented by \textsc{MPNet-1B}, a sentence model trained on over 1 billion text pairs\footnote{\textsc{MPNet-1B}: \url{https://bit.ly/2Zbm2Iq}}, Sentence-Bert (\textsc{SentBert})  \cite{reimers2019sentencebert}, \textsc{SimCSE} \cite{gao2021simcse}, \texttt{cosentbert} of \S\ref{sec-single-match}, and \textsc{ICTSentBert} \cite{lee2019latent}.\\
    \item Abstract models: The abstract level model \textsc{Specter} \cite{cohan2020specter}, represents a SOTA model for scientific document similarity  trained on \emph{cited} abstract pairs. We also train a variant of this model on \emph{co-cited} papers: \textsc{Specter-CoCite}. Finally, we also compare to \textsc{SciNCL}, introduced in recent concurrent work of \citet{ostendorff2021scincl}. \textsc{SciNCL} presents a bi-encoder model similar to \textsc{Specter} with improvements to its contrastive learning procedure -- presenting a complementary direction to our approach.\\
    \item Sentence level models modified for whole abstract similarity: Here we combine the SOTA sentence encoder \textsc{MPNet-1B} with the optimal transport (\S\ref{sec-multi-align-ot}) for aggregating sentence level matches giving \textsc{otMPNet-1B}.
\end{enumerate*}

Sentence models use the same inference procedure as our single match method, abstract models rank using L2 distances between papers embeddings, and the modified sentence model uses the multi match inference procedure. All reported model hyper-parameters are tuned, trained on 1.3M co-citation triples, and initialized with \textsc{Specter} unless noted otherwise.\footnote{Initialization indicated via subscript in tables.} In reporting results, we report standard retrieval metrics Mean Average Precision (MAP) and NDCG at rank K. For NDCG@K we follow \citet{wang13ndcg}, and choose K$=p*|\mathcal{C}|$ where $p\in(0,1)$. NDCG$_{\%20}$ therefore refers to NDCG computed at 20\% of the pool size for a query. This is apt since queries have varying pool sizes.
Appendices \ref{appendix-cocite-data}, \ref{appendix-training}, and \ref{appendix-hyper-params} detail training data, algorithms, and hyper-parameters. Next, we present our main results comparing proposed approaches to baselines.

\subsection{Results}
\label{sec-main-results}
\textbf{Fine-grained similarity}: Table \ref{tab-model-results-compsci} presents results 
on \textsc{CSFCube}. We report 
performance on the three facets \texttt{background}, \texttt{method}, and \texttt{result} annotated in the dataset, and aggregated across all facets. 
We first make some observations about baseline methods:
\begin{enumerate*}
    \item  \textsc{MPNet-1B} outperforms all other sentence level models and a SOTA abstract representation, \textsc{specter}, indicating the value of sentence-level information for capturing fine-grained similarities. With \textsc{otMPNet-1B} indicating the value of modeling multiple matches.%
    \item  \textsc{Specter-CoCite}$_\mathtt{Scib}$, which is identical to  \textsc{Specter} but trained on co-citations outperforms it, showing the value of co-citations for fine-grained similarity.%
\end{enumerate*}

Next, we examine performance of the proposed methods:
\begin{enumerate*}
    \item First we note that all of the proposed approaches consistently outperform performant prior work,  \textsc{ot}/\textsc{MPNet-1B} and  \textsc{Specter}, by about 5-6 points, and concurrent work of \textsc{SciNCL} by  about 1.5-2 points aggregated across queries.%
    \item Next, we note that the proposed approaches outperform  \textsc{Specter-CoCite$_\mathtt{Spec}$}, trained on co-citations by 2-3 points aggregated across queries.%
    \item Our single match model trained with textual supervision, \textsc{ts\aspect} consistently outperforms baselines. 
    \item Finally, our multi-match model \textsc{ot\aspect}, while outperforming baselines sees aggregate performance similar to single match methods. This is reasonable given the aspect-specific annotation of \textsc{CSFCube} where we expect gains from modeling fine-grained (contextualized) matches rather than aggregating multiple matches.
\end{enumerate*} 

Now, we examine facet-specific performance:
\begin{enumerate*}
    \item Performance on \texttt{background} sees higher performance in general and the smallest gains for the proposed approaches. This may be attributed to \texttt{background} similarity being captured in \emph{coarse}-grained topical similarity, a trait largely captured in existing baselines.
    \item \texttt{method} similarity in \textsc{CSFCube} presents significant challenges \cite[Sec 6]{MysoreCSFCUBE2021} since it relies upon procedural similarities across steps of a method and on domain knowledge based similarities - this is often captured in co-citation data (Fig \ref{fig-text-example} presents one such complex paraphrase example). We see strongest performance for \textsc{ts\aspect} here.
    \item Finally, given that paper results interpretations are often dependent on all aspects of a given paper, \texttt{result} similarity often depends on similarity across the whole abstract. This leads \textsc{ot\aspect} which models multiple matches to see strong performance.
\end{enumerate*}

\textbf{Whole-abstract similarity:} Table \ref{tab-model-results-biomed} presents 
results on \textsc{treccovid}$_\mathtt{RF}$ and \textsc{relish}. At the outset, we note 
that while being annotated for whole-abstract relevance, these datasets present 
different characteristics. While \textsc{treccovid}$_\mathtt{RF}$ presents 
queries centered on a very specific topic, \textsc{relish} presents a much more 
diverse set of queries. Further, \textsc{treccovid}$_\mathtt{RF}$ pairs queries 
with pools of about 9000 candidates while \textsc{relish} has about 60 
candidates per query. Next, we examine baselines.

\begin{enumerate*}
    \item In contrast to fine-grained similarity datasets the best sentence level model  \textsc{MPNet-1B}, significantly underperforms an abstract level model,  \textsc{Specter}, indicating the need for whole abstract representations for these datasets. Aggregating sentence matches as in \textsc{otMPNet-1B}, drastically improves \textsc{MPNet-1B}.
    \item Next, similar to results in Table \ref{tab-model-results-compsci}, a model identical to  \textsc{Specter}, but trained on co-citations,  \textsc{Specter-CoCite$_\mathtt{Spec}$}, outperforms  \textsc{Specter} indicating the value of co-citation signal for whole-abstract similarity too.
    \item Finally, we also note that while \textsc{SciNCL} sees an expected stronger performance to \textsc{Specter} in \textsc{relish}, it sees comparable performance in \textsc{treccovid}$_\mathtt{RF}$ -- indicating the influence of the candidate pool size on its performance.
\end{enumerate*}

In examining performance of our proposed methods, we note the following:
\begin{enumerate*}
    \item Across datasets, our method for single matches, \textsc{ts\aspect}, outperforms context-independent sentence baselines by several points indicating the value of contextualization. However, this method still underperforms abstract-level baselines.
    \item However, methods modeling multiple matches, \textsc{ot\aspect} and \textsc{ts+ot\aspect}, substantially outperform \textsc{ts\aspect} as well as baseline prior work \textsc{Specter} and \textsc{otMPNet-1B}. This performance indicates the strength of OT based aggregation of fine-grained matches for abstract level similarity. The proposed methods additionally match or outperform the concurrent approaches of \textsc{SciNCL}. Note here, that given the complementary approach presented in \textsc{SciNCL} - strong models are likely to result from combining both approaches.
\end{enumerate*}

We present results demonstrating the value of the proposed approach on the \textsc{SciDocs} benchmark in Appendix \ref{appendix-scidocs}. Further, we also present a set of ablations in Appendix \ref{appendix-ablations}. These ablations establish the value of textual supervision over the encoder ($\mathtt{BERT}_{\mathcal{E}}$) used for encoding the text, the value of optimal transport compared to attention alternatives, and alternative single-match models trained without co-citation contexts.

\section{Related Work}

\textit{Aspect-based paper representations:} 
A large body of work learns structured representations of scientific papers. \citet{jain2018learning} present an approach which learns pre-defined aspect (PICO) encoders for biomedical papers.
Similarly work of \citet{neves2019evaluation}, \citet{chan2018solvent}, and \citet{kobayashi2018citation} each label paper texts and then compute aspect-specific embeddings for document classification or ranking using existing methods. This line of work often relies on pre-defined aspects and building aspect-specific methods. Finally, work of \citet{ostendorff2020aspectcoling} and \citet{luu2021explaining} present an approach with some similarities to the ones presented above -- these approaches leverage classification or language generation models to output fine-grained relationships between pairs of papers. Our work leverages co-citation contexts to supervise free-text aspects with a new model for document retrieval, that is also not tied to a specific schema of labels.

\textit{Fine-grained document representations:} Another similar line of work is modeling fine-grained document-document similarity at the level of words or latent topics. 
Examples include early work \citet{el2011qbepapers} presenting paper recommendation methods with unigram-level similarity between papers using authorship and citation links or using latent document topics \cite{gong2018documenttopics,Yurochkin2019topicOT,dieng2020ETM}.

Our approach represents documents via sentences, a common and intuitive structure for reasoning about scientific document facets \cite{chan2018solvent,zhou2020multilevel}. 
\citet{ginzburg2021self} present a self-supervised model for contextual sentence representations in long documents similar to our ICT baseline \cite{lee2019latent}. 

\textit{Ad-hoc Search:} 
A range of recent work in information retrieval presents multi-vector models intended to capture different aspects of candidate documents with score aggregation relying on summations, max, or attention functions \cite{khattab2020colbert, luan2021sparse, Humeau2020Polyencoders}, these however focusing on short-text queries seen in search or question answering (QA). \citet{Mitra2017localdistr} explore an approach to model term-level fine-grained similarities with neural networks, \citet{liu2018entityduet} model fine-grained matches at the level of entity spans, and \citet{Yilmaz2019cross} model document relevance by aggregating sentence relevance. Similarly, recent work of \citet{lee2021densephrases} models fine-grained matches for QA at the phrase level. Importantly, these methods rely on supervision from knowledge bases or QA datasets, limiting applicability to specific span definitions and areas with these resources, often not present in the scientific domain \cite{hope-etal-2021-extracting}.

A range of modeling approaches in the context of other tasks resemble elements of our approach. We describe these in Appendix \ref{appendix-relwork}.
\section{Conclusions}
\label{sec-conclusions}

We presented \aspect, a scientific document similarity model that is trained by leveraging co-citation contexts for learning fine-grained similarity.  We use co-citation contexts as a novel form of \emph{textual} supervision to guide the learning of multi-vector document representations. Our model outperformed strong baselines on seven document similarity tasks across four English scientific text datasets. Moreover, we showed that a fast single-match method achieves competitive results, enabling fine-grained document similarity in large-scale scientific corpora. A future direction is the interactive use of our methods, with a system allowing users to highlight specific aspects of papers and retrieve contextually-relevant matches. Another promising application is for finding analogies --- structural matches between texts describing ideas, as in scientific papers, to boost discovery \cite{hope2017accelerating,hope2021scaling,chan2018solvent}.

\section{Acknowledgements}
\label{sec-acks}
We are grateful to the members and fellow interns of Semantic Scholar at AI2, and members of the IESL and CIIR labs at the University of Massachusetts Amherst for helpful discussions. This work was supported in part by the National Science Foundation under Grant Number IIS-1922090, the Chan Zuckerberg Initiative under the project Scientific Knowledge Base Construction, the NSF Convergence Accelerator Award \#2132318, and using high performance computing equipment obtained under a grant from the Collaborative R\&D Fund managed by the Massachusetts Technology Collaborative.


\bibliography{facetdiscovery-apps}
\bibliographystyle{acl_natbib}

\appendix
\section{Co-citation Data}
\label{appendix-cocite-data}
As noted in \S\ref{sec-fine-grained-data}, we train the proposed 
methods on English co-cited papers. 
We build a dataset of co-cited papers from the S2ORC corpus\footnote{Released under a CC BY-NC 2.0. license.} \cite{lo2020s2orc}. Since our evaluation datasets draw on text from different domains we build training 
sets with co-cited papers for each: biomedicine for \textsc{relish} and 
\textsc{treccovid}$_{\mathtt{RF}}$, computer science for \textsc{CSFCube}, and a 60/40 mix of biomedicine and CS for \textsc{SciDocs}. Each dataset 
contains 1.3M training triples. 

Next we describe construction of our co-citation data given 8.1 million English full text articles in the S2ORC corpus which have been parsed for citation mentions and linked to cited papers in the corpus using automatic tools \cite{lo2020s2orc}:
\begin{itemize}[noitemsep, nolistsep]
    \item Domain definition: We define our biomedical articles to be those tagged either ``Medicine'' or ``Biology'' in S2ORC. ``Computer Science'' tagged papers are treated as CS papers.
    \item Co-citation contexts: To obtain co-cotation contexts - we first obtain sentence boundaries for co-citation contexts using the \texttt{en\_core\_sci\_sm} pipeline included in \texttt{spacy}.\footnote{\url{https://allenai.github.io/scispacy/}}. 
    \item Filtering abstracts: In selecting abstracts for our dataset we retain those that have a minimum of 3 sentences, and a maximum of 20 sentences. Further, abstracts where all the sentences are too small (3 tokens) are excluded. Similarly, abstracts with sentences greater than 80 tokens are excluded.
    \item Selecting training co-citated abstract data $\{\mathcal{P}\}$: Given contexts with qualifying abstracts as described above, we only retain co-citation contexts with 2 or 3 co-cited papers. A manual examination revealed that larger co-cited sets tended to be more loosely related. 
    \item Selecting co-citation sentence training data for $\mathtt{BERT}_{\mathcal{E}}$: Note that this represents a sentence encoder trained by treating co-citation contexts referencing the same paper as paraphrases. Here, we select co-citation contexts containing 2 or more co-cited papers as paraphrase sets $\mathcal{E}$.
\end{itemize}
Abstract level training triples for the biomedical and computer science sets are built by treating all unique pairs of papers as positives. 1.3 million triples were used for each domain - these were sampled from larger sets at random. 

\section{Evaluation Dataset Details}
\label{appendix-evaldata}
Here we provide further detail on the evaluation datasets overviewed in \S\ref{sec-experiments}.

\textsc{relish}: An annotated dataset of biomedical abstract queries labelled by experts \cite{Brown2019RELISH}. In a number of cases expert annotators are the authors of query papers. Per query candidate pools are of size 60, with 1638 queries in development and test sets each. Dataset is released under a Creative Commons Attribution 4.0 International License.

\textsc{treccovid}$_{\mathtt{RF}}$: While the original \textsc{treccovid} dataset of \citet{treccovid2021} is labelled for ad-hoc search by experts, we reformulate the dataset for abstract similarity, treating all documents relevant to one ad-hoc query as similar to each other. From each original query and its respective relevance-labeled documents, we sample an abstract from relevant documents (relevance of 2) and use that as our query document. We treat all other relevant documents as positive examples for the query. Documents relevant for \emph{other} queries are treated as irrelevant for the sampled query. This results in about 9000 candidates per query abstract in \textsc{treccovid}$_{\mathtt{RF}}$. \textsc{treccovid}$_{\mathtt{RF}}$ consists of about 1200 queries in the development and test splits each. This dataset builds on the CORD-19 dataset \cite{wang2020cord} released under a Apache License 2.0, the license of \textsc{treccovid} however isn't clear from the dataset release.

\textsc{SciDocs}: A benchmark suite of tasks intended for abstract-level scientific document representations \cite{cohan2020specter}. We evaluate our methods on the tasks of predicting: citations, co-citations, co-views, and co-reads. Per query candidate pools are of size 30 about 1000 queries per task and development and test split. We exclude classification and recommendation sub-tasks relying on additional inference components. Dataset is released under a GNU General Public License v3.0 license.

\textsc{CSFCube}: The dataset consists of 50 queries labelled for relevance against about 120 candidates per query. Dataset is released under a Creative Commons Attribution-NonCommercial 4.0 International license.

\section{Co-citation Context Encoder}
\label{appendix-cosentbert}
Here we present details of alternative design choices for our co-citation context encoder. In the use of $\mathtt{BERT}_{\mathcal{E}}$, we note in \S\ref{sec-single-match} that this encoder is kept frozen during the course of training $\mathtt{BERT}_{\theta}$. Fine-tuning $\mathtt{BERT}_{\mathcal{E}}$ via a straight-through estimator \cite{bengio2013estimating} under-performed freezing it. Using other encoders for scientific text such as \textsc{specter} as $\mathtt{BERT}_{\mathcal{E}}$ under-performed \texttt{CoSentBert}. A recent strong model for sentence representation \texttt{MPNet-1B}\footnote{\texttt{MPNet-1B}: \url{https://bit.ly/2Zbm2Iq}} lead to similar performance on abstract and aspect-conditional tasks as \texttt{CoSentBert}, indicating that a minimum requisite sentence encoder is all that is needed for $\mathtt{BERT}_{\mathcal{E}}$.

\section{Baselines}
\label{appendix-baselines}
Here we provide further detail on the baselines overviewed in \S\ref{sec-experiments}.

\begin{description}[noitemsep, nolistsep]
    \item \textsc{MPNet-1B} \& \textsc{otMPNet-1B}: A sentence level baseline of a MPNet \cite{song2020mpnet} base model, fine-tuned on 1.17 billion similar text pairs in a contrastive learning setup.\footnote{\texttt{MPNet-1B}: \url{https://bit.ly/2Zbm2Iq}} This training data broadly represents web and scientific texts. Further we combine \textsc{MPNet-1B} with an OT based aggregation scheme similar to our multi-match model to yield, \textsc{otMPNet-1B} a baseline using optimal transport with a performant sentence encoder.
    \item SimCSE: A recent sentence representation model \cite{gao2021simcse}. We compare to two model variants: an unsupervised model \textsc{UnSimCSE-BERT}, and a variant supervised with NLI data, \textsc{SuSimCSE-BERT}.
    \item Sentence-Bert: A sentence level transformer model fine-tuned on similar sentence pairs \cite{reimers2019sentencebert}. We compare performance to two variants, \textsc{SentBert-PP} and \textsc{SentBert-NLI}, fine-tuned on paraphrases and natural language inference (NLI) data respectively.
    \item \texttt{CoSentBert}: The sentence-level model we describe in \S\ref{sec-single-match}: A \textsc{SciBERT} model fine-tuned on co-citation sentence contexts referencing the same set of co-cited papers.
    \item \textsc{ICTSentBert}: A \textsc{SciBERT} sentence model trained using the self-supervised inverse close task \cite{lee2019latent}. Here we train abstract sentence representations to capture the semantics of their paragraph contexts.
    \item \textsc{Specter}: A state of the art abstract level representation \cite{cohan2020specter}. Here a \textsc{SciBERT} model is fine-tuned to maximize similarity between representations of \emph{cited} papers. We also train a variant of this model on \emph{co-cited} papers: \textsc{Specter-CoCite}.
    \item \textsc{SciNCL}: A recent concurrent state of the art abstract level representation \cite{ostendorff2021scincl}. This approach trains a model similar to \textsc{Specter}, with improvements to the negative sampling strategies of \citet{cohan2020specter} for contrastive learning. This presents a complementary contribution to the one presented in our work - with future modeling approaches likely to benefit from both approaches.
\end{description}

For the baselines described above specific model names from the Hugging Face\footnote{\url{https://huggingface.co/}} and Sentence Transformers\footnote{\url{https://www.sbert.net/docs/pretrained_models.html}} libraries are as follows:
\begin{description}[noitemsep, nolistsep]
    \item {MPNet-1B}: HF; {sentence-transformers/all-mpnet-base-v2}.
    \item SimCSE: HF; {princeton-nlp/sup-simcse-bert-base-uncased}, {princeton-nlp/unsup-simcse-bert-base-uncased}.
    \item Sentence-Bert: ST; Paraphrases: {paraphrase-TinyBERT-L6-v2}. NLI: {nli-roberta-base-v2} from the Sentence-Transformers library.
\end{description}

\section{Training}
\label{appendix-training}
All our approaches are trained using the  Adam optimizer with an initial linear warm-up for 2000 steps followed by a linear decay using gradient accumilation for a batch size of 30. The 
margin $m$ in the triplet loss is set to 1. We implement all methods using PyTorch, HuggingFace, and GeomLoss libraries. Training convergence is established based on the loss on a held out set of co-citation data ensuring that training does not rely on a labelled dataset for convergence checks.

All experiments were run with data parallelism over servers nodes with the following GPU configurations: 8$\times$12GB NVIDIA GeForce GTX 1080 Ti GPUs, 4$\times$24GB NVIDIA Tesla M40 GPUs, or 2$\times$48GB NVIDIA Quadro RTX 8000 GPUs. Servers had 12-24 CPUs per node and 256-385GB RAM. The training time per experiment varied from 5-20 hours, and the experiments in this paper represent about 4746 GPU hours of training.

\section{Model Hyper-Parameters}
\label{appendix-hyper-params}
Here we report the best performing model hyper-parameters. This is done per training dataset. For computer science trained models evaluated on \textsc{CSFCube}:
\begin{itemize}[noitemsep]
    \item Specter-CoCite$_\mathtt{Scib}$: LR 2e-5. 
    \item Specter-CoCite$_\mathtt{Spec}$: LR 2e-5.
    \item \textsc{ts\aspect}$_\mathtt{Spec}$: LR 2e-5. 
    \item \textsc{ot\aspect}$_\mathtt{Spec}$: LR 2e-5. $\tau$ 0.5.
    \item \textsc{ts+ot\aspect}$_\mathtt{Spec}$: LR 1e-5.  $\tau$ 0.5.
\end{itemize}

\begin{table*}[t]
\centering
\footnotesize
\begin{tabular}{@{}lcc cccccc@{}}
\toprule
SciDocs tasks $\rightarrow$ & 
\multicolumn{2}{c}{\texttt{Citations}}   &
\multicolumn{2}{c}{\texttt{Co-Citations}}       & 
\multicolumn{2}{c}{\texttt{Co-Reads}}       &
\multicolumn{2}{c}{\texttt{Co-Views}}           \\ \cmidrule(lr){2-3} \cmidrule(lr){4-5} \cmidrule(lr){6-7} \cmidrule(lr){8-9}
Models & MAP   & NDCG     &  MAP  & NDCG & MAP   & NDCG     &  MAP  & NDCG  \\\toprule
\textsc{MPNet-1B}            & 86.76 & 92.63 & 85.68 & 92.16 & 83.45 & 90.47 & 82.51 & 89.29\\\midrule
\textsc{Specter}                & \textbf{92.39} & \textbf{95.90} & 88.32 & 93.88 & 86.42 & 92.39 & 84.65 & 90.70\\
\textsc{Specter-CoCite$_\mathtt{Scib}$}   & $\underset{\pm 0.33}{89.16}$ & $\underset{\pm 0.28}{93.97}$ & $\underset{\pm 0.18}{90.21}$ & $\underset{\pm 0.14}{94.76}$ & $\underset{\pm 0.22}{86.85}$ & $\underset{\pm 0.18}{92.51}$ & $\underset{\pm 0.16}{85.70}$ & $\underset{\pm 0.09}{91.37}$\\
\textsc{Specter-CoCite$_\mathtt{Spec}$}   & $\underset{\pm 0.10}{89.85}$ & $\underset{\pm 0.08}{94.26}$ & $\underset{\pm 0.17}{90.82}$ & $\underset{\pm 0.11}{95.11}$ & $\underset{\pm 0.14}{87.14}$ & $\underset{\pm 0.13}{92.65}$ & $\underset{\pm 0.10}{85.81}$ & $\underset{\pm 0.05}{91.35}$\\\midrule
\textsc{ts\aspect}$_\mathtt{Spec}$             & $\underset{\pm 0.26}{90.99}$ & $\underset{\pm 0.17}{95.04}$ & ${\underset{\pm 0.06}{\bf 90.92}}$ & ${\underset{\pm 0.05}{\bf 95.26}}$ & $\underset{\pm 0.07}{87.51}$ & $\underset{\pm 0.06}{92.97}$ & ${\underset{\pm 0.20}{\bf 85.87}}$ & ${\underset{\pm 0.14}{\bf 91.46}}$\\
\textsc{ot\aspect}$_\mathtt{Spec}$          & $\underset{\pm 0.28}{91.13}$ & $\underset{\pm 0.20}{95.08}$ & $\underset{\pm 0.13}{90.88}$ & $\underset{\pm 0.02}{95.25}$ & $\underset{\pm 0.14}{87.50}$ & $\underset{\pm 0.12}{92.90}$ & $\underset{\pm 0.20}{85.70}$ & $\underset{\pm 0.11}{91.30}$\\
\textsc{ts+ot\aspect}$_\mathtt{Spec}$     & $\underset{\pm 0.33}{91.09}$ & $\underset{\pm 0.17}{95.03}$ & $\underset{\pm 0.08}{90.83}$ & $\underset{\pm 0.05}{95.22}$ & ${\underset{\pm 0.05}{\bf 87.60}}$ & ${\underset{\pm 0.01}{\bf 92.98}}$ & $\underset{\pm 0.25}{85.81}$ & $\underset{\pm 0.15}{91.42}$\\\bottomrule
\end{tabular}

\caption{Test set results for baseline and proposed methods on sub-tasks included in the \textsc{SciDocs} benchmark. Our approaches outperform a prior strong model,  \textsc{Specter}, by 1-1.5 points on 3 of 4 sub-tasks. Metrics (MAP, NDCG) are computed based on averages over three re-runs of models.  \textsc{Specter} uses model parameters as part of the Huggingface library. Here, \textsc{ts\aspect}: Text supervised single-match method, \textsc{ot\aspect}: Optimal Transport multi-match method and \textsc{ts+ot\aspect}: Multi-task multi aspect method.}
\label{tab-model-results-scidocs-full}
\end{table*}

For biomedical trained models evaluated on \textsc{treccovid} and \textsc{relish}:
\begin{itemize}[noitemsep]
    \item Specter-CoCite$_\mathtt{Scib}$: LR 2e-5. 
    \item Specter-CoCite$_\mathtt{Spec}$: LR 2e-5.
    \item \textsc{ts\aspect}$_\mathtt{Spec}$: LR 2e-5. 
    \item \textsc{ot\aspect}$_\mathtt{Spec}$: LR 2e-5. $\tau$ $5000$.
    \item \textsc{ts+ot\aspect}$_\mathtt{Spec}$: LR 2e-5.  $\tau$ $5000$.
\end{itemize}

For biomedical+computer science trained models evaluated on \textsc{treccovid} and \textsc{relish}:
\begin{itemize}[noitemsep]
    \item Specter-CoCite$_\mathtt{Scib}$: LR 2e-5. 
    \item Specter-CoCite$_\mathtt{Spec}$: LR 2e-5.
    \item \textsc{ts\aspect}$_\mathtt{Spec}$: LR 1e-5.
    \item \textsc{ot\aspect}$_\mathtt{Spec}$: LR 1e-5. $\tau$ $5000$.
    \item \textsc{ts+ot\aspect}$_\mathtt{Spec}$: LR 1e-5.  $\tau$ $5000$.
\end{itemize}
We found it beneficial to use a low temperature $\tau$ in computing distributions $\mathbf{x}$ for OT computation for \textsc{CSFCube} - a fine-grained similarity dataset. On the other hand we found it beneficial to use a high temperature $\tau$ in computing distributions $\mathbf{x}$, causing it to be effectively uniform, for OT computation in whole-abstract datasets \textsc{SciDocs}, \textsc{relish}, and \textsc{treccovid}$_\mathtt{RF}$. This is reasonable given the nature of similarity captured in these datasets. Hyper-parameters of the underlying encoders were not changed from their default values -- other hyper-parameters are common to methods and desribed in \S\ref{sec-experiments}.

Finally, in computing OT transport plans, we optimize a entropy regularized objective: $\underset{\mathbf{P} \in \mathcal{S}}{\mathtt{min}}\langle\mathbf{D},\mathbf{P}\rangle - \frac{1}{\lambda}h(\mathbf{P})$. Our experiments use a fixed value of $\lambda=20$. 

\textbf{Hyper-parameter tuning:} We tune the hyper-parameters of all the ablated and proposed methods across the different datasets on development set performance. For \textsc{CSFCube} the \textit{Aggregated} dev set performance was used for computer science training data models, \textsc{treccovid}$_\mathtt{RF}$ and \textsc{relish} dev sets were used for biomedical data models with ties between the two broken by the more challenging \textsc{treccovid}$_\mathtt{RF}$ performance, and computer science +biomedical data models were tuned on average task performance of \textsc{SciDocs} tasks. Given the expense of training models (about 20h for the proposed models) we first tune softmax temperatures then tuned learning rates. Large changes across learning rates weren't observed for the models.
All learning rates are tuned over the range \{1e-5, 2e-5, 3e-5\}, OT sentence softmax temperatures $\tau$ are tuned over \{0.5, 1, 5, 5000\}, and softmax temperatures for ablation A3 was tuned over \{0.5, 1, 5\}.

\section{\textsc{SciDocs} Benchmark Result}
\label{appendix-scidocs}
\textbf{SciDocs Benchmark:} Table \ref{tab-model-results-scidocs-full} indicates performance on the abstract level document similarity benchmark \textsc{SciDocs} of \citet{cohan2020specter}. First we note that the strong performance of  \textsc{Specter} indicates a smaller gap to be closed. Here, although our proposed methods see similar performance to each other they consistently outperform  \textsc{Specter} on 3 of 4 tasks establishing state of the art performance. Given  \textsc{Specter}'s citation training signal and our co-citation signal, we see  better performance on the \texttt{Citations} and \texttt{Co-Citation} tasks respectively. Finally, note that our co-citation trained approaches broadly see better performance (1-1.5 points) on extrinsic tasks of \texttt{Co-Reads} and \texttt{Co-Views} indicating the value of this signal.

\section{Ablations}
Here we ablate a range of model components in establishing factors which contribute performance. In ablations we only report performance on \textsc{CSFCube}, \textsc{treccovid}$_{\mathtt{RF}}$, and \textsc{relish}.

\label{appendix-ablations}
\begin{table}[t]
\centering
\footnotesize

\begin{tabular}{@{}lrr@{}} 
 \toprule
 \textsc{CSFCube} \textit{Agg.} & MAP   & NDCG$_{\%20}$ \\\midrule
\textsc{abs\aspect}$_\mathtt{Spec}$             & $\underset{\pm 1.39}{37.03}$ & $\underset{\pm 0.76}{59.57}$\\
\textsc{ts\aspect}$_\mathtt{Spec}$             & $\underset{\pm 0.93}{40.26}$ & $\underset{\pm 0.67}{60.71}$\\\bottomrule
\end{tabular}\\[4pt]
\vspace{10pt}
\setlength{\tabcolsep}{3pt}
\begin{tabular}{@{}lrrrr@{}}
 & \multicolumn{2}{c}{\textsc{treccovid}$_\mathtt{RF}$} 
 & \multicolumn{2}{c}{\textsc{relish}}\\
 \cmidrule(lr){2-3} \cmidrule(lr){4-5} 
 &  MAP  & NDCG$_{\%20}$ & MAP  & NDCG$_{\%20}$  \\\midrule
\textsc{abs\aspect}$_\mathtt{Spec}$      & $\underset{\pm 0.9}{25.42}$ & $\underset{\pm 0.55}{55.34}$ & $\underset{\pm 0.69}{58.78}$ & $\underset{\pm 0.57}{75.80}$\\
\textsc{ts\aspect}$_\mathtt{Spec}$             & $\underset{\pm 0.45}{26.24}$ & $\underset{\pm 0.65}{56.55}$ & $\underset{\pm 0.51}{61.29}$ & $\underset{\pm 0.42}{ 77.89}$\\\bottomrule
\end{tabular}
\caption{Results for Ablation A1. Performance of \textsc{ts\aspect} trained with textual supervision from co-citation contexts ablated for the effect of the text vs. influence of the text encoder ($\mathtt{BERT}_{\mathcal{E}}$=\texttt{CoSentBert}; in \S\ref{sec-single-match}) used to compute alignments to the co-citation contexts. Standard deviation across 3 model re-runs under mean performance.}
\label{tab-finegrained-absali}
\end{table}
\textbf{A1.\ Does \textsc{ts\aspect} gain from textual supervision over the encoder used to compute alignment?} \textsc{ts\aspect} relies upon a sentence alignment encoder, $\mathtt{BERT}_{\mathcal{E}}$ in \S\ref{sec-single-match}, to 
compute alignments, $\hat{i_p}, \hat{i}_{p^{'}}$, from the co-citation context 
to the co-cited abstracts. Here we investigate if improvements in 
\textsc{ts\aspect} are attributable to $\mathtt{BERT}_{\mathcal{E}}$ or to the co-
citation contexts themselves. We investigate this by comparing the performance 
of \textsc{ts\aspect} to a model trained to maximize the alignment between 
abstract sentences directly computed using $\mathtt{BERT}_{\mathcal{E}}$, we 
refer to this as \textsc{abs\aspect}. This may be viewed as a form of 
knowledge distillation where alignments from a more local sentence encoder model, $\mathtt{BERT}_{\mathcal{E}}$, are 
distilled into the contextual sentence encoder of \textsc{ts\aspect}. As Table \ref{tab-finegrained-absali} 
shows, \textsc{ts\aspect} consistently outperforms \textsc{abs\aspect}, 
indicating the value added by natural language supervision from the co-citation 
contexts.

\begin{table}[t]
\centering
\footnotesize
\begin{tabular}{@{}lrr@{}} \toprule
 \textsc{CSFCube} \textit{Agg.} & MAP   & NDCG$_{\%20}$ \\\midrule
\textsc{att\aspect}$_\mathtt{Spec}$             & $\underset{\pm 1.52}{41.85}$ & $\underset{\pm 0.82}{61.67}$\\
\textsc{ot\aspect}$_\mathtt{Spec}$             & $\underset{\pm 0.53}{ 40.79}$ & $\underset{\pm 0.52}{61.41}$\\\bottomrule
\end{tabular}\\[4pt]
\vspace{10pt}
\setlength{\tabcolsep}{3pt}
\begin{tabular}{@{}lrrrr@{}}
 & \multicolumn{2}{c}{\textsc{treccovid}$_\mathtt{RF}$} 
 & \multicolumn{2}{c}{\textsc{relish}}\\
 \cmidrule(lr){2-3} \cmidrule(lr){4-5} 
 &  MAP  & NDCG$_{\%20}$ & MAP  & NDCG$_{\%20}$  \\\midrule
\textsc{att\aspect}$_\mathtt{Spec}$   & $\underset{\pm 0.78}{29.51}$ & $\underset{\pm 0.51}{60.96}$ & $\underset{\pm 0.52}{61.92}$ & $\underset{\pm 0.50}{78.54}$\\
\textsc{ot\aspect}$_\mathtt{Spec}$    & $\underset{\pm 0.53}{30.92}$ & $\underset{\pm 0.67}{62.23}$ & $\underset{\pm 0.29}{62.57}$ & $\underset{\pm 0.26}{78.95}$\\\bottomrule
\end{tabular}
\caption{Results for Ablation A2. Performance for an alternative method, \textsc{att\aspect}, for modeling multiple matches with an attention mechanism instead of optimal transport in the proposed method. Standard deviation across 3 model re-runs under mean performance.}
\label{tab-attention-align}
\end{table}
\textbf{A2.\ Can multi-aspect matching use attention aggregation instead of optimal transport?} Since our multi-aspect match model uses a soft sparse matching with optimal transport we examine contributions of this component by comparing performance of a model (\textsc{att\aspect}) trained with soft-alignment using an attention mask, $\mathbf{A}$ -- attention is also a popular choice in prior work \citet{Humeau2020Polyencoders, zhou2020multilevel}. Here, $f_{\mathtt{Att}}(p,p') = \langle\mathbf{D}, {\mathbf{A}}\rangle$ with, $\mathbf{A} = \mathtt{softmax}(-\mathbf{D}/\tau)$. Note that OT imposes specific inductive bias via the structure of the trasport plan in ensuring it to be a permutation matrix - a desirable property in computing multiple alignments between a set of points. Table \ref{tab-attention-align} examines performance of these model variants. Broadly, \textsc{att\aspect} sees performance comparable or worse than  \textsc{ot\aspect}. While \textsc{att\aspect} sees improved performance in \textsc{CSFCube} it sees much larger variation across runs. In our abstract retrieval datasets, where we expect gains from modeling multiple matches, we see better or similar performance from \textsc{ot\aspect} over \textsc{att\aspect}.

\begin{table}[t]
\centering
\footnotesize
\begin{tabular}{@{}lrr@{}}
 \toprule 
 \textsc{CSFCube} \textit{Aggregated} & \textsc{map}   & \textsc{ndcg}$_{\%20}$  \\\toprule
\textsc{max\aspect}$_\mathtt{SciB}$        & $\underset{\pm 1.37}{36.66}$ & $\underset{\pm 0.86}{57.68}$\\
\textsc{max\aspect}$_\mathtt{Spec}$        & $\underset{\pm 1.38}{ 39.42}$ & $\underset{\pm 1.53}{ 60.63}$\\\midrule
\textsc{ts\aspect}$_\mathtt{SciB}$        & $\underset{\pm 0.76}{40.10}$ & $\underset{\pm 0.61}{60.92}$\\
\textsc{ts\aspect}$_\mathtt{Spec}$        & $\underset{\pm 0.93}{40.26}$ & $\underset{\pm 0.67}{60.71}$ \\\bottomrule
\end{tabular}\\[4pt]
\vspace{10pt}
\setlength{\tabcolsep}{4pt}
\begin{tabular}{@{}lrrrr@{}}  
 & \multicolumn{2}{c}{\textsc{treccovid}$_\mathtt{RF}$} 
 & \multicolumn{2}{c}{\textsc{relish}}\\ \cmidrule(lr){2-3} \cmidrule(lr){4-5} 
 &  \textsc{map}  & \textsc{ndcg}$_{\%20}$ & \textsc{map}  & \textsc{ndcg}$_{\%20}$  \\\midrule
\textsc{max\aspect}$_\mathtt{SciB}$        & $\underset{\pm 1.15}{24.87}$ & $\underset{\pm 1.49}{54.33}$ & $\underset{\pm 0.31}{61.36}$ & $\underset{\pm 0.24}{78.10}$\\
\textsc{max\aspect}$_\mathtt{Spec}$        & $\underset{\pm 0.85}{ 25.84}$ & $\underset{\pm 1.21}{ 56.52}$ & $\underset{\pm 0.97}{61.20}$ & $\underset{\pm 0.36}{78.00}$\\\midrule
\textsc{ts\aspect}$_\mathtt{SciB}$             & $\underset{\pm 0.71}{27.68}$ & $\underset{\pm 0.75}{58.42}$ & $\underset{\pm 0.31}{61.45}$ & $\underset{\pm 0.33}{78.12}$\\
\textsc{ts\aspect}$_\mathtt{Spec}$             & $\underset{\pm 0.45}{26.24}$ & $\underset{\pm 0.65}{56.55}$ & $\underset{\pm 0.51}{61.29}$ & $\underset{\pm 0.42}{77.89}$\\\bottomrule
\end{tabular}
\caption{Results for Ablation A3. Performance of a simpler single-match model, \textsc{max\aspect}, trained using only $\mathtt{BERT}_{\theta}$ representations while also varying encoder initialization between \textsc{Specter} and \textsc{SciBert} (indicated as subscripts for models). Standard deviation across 3 model re-runs under mean performance.}
\label{tab-scib-spec-finegrained}
\end{table}

\begin{table*}[t]
\centering
\footnotesize
\begin{tabular}{@{}lcc cccccc@{}} \toprule
\textsc{CSFCube} facets$\rightarrow$ & 
\multicolumn{2}{c}{\textit{Aggregated}}   &
\multicolumn{2}{c}{\texttt{background}}       & 
\multicolumn{2}{c}{\texttt{method}}       &
\multicolumn{2}{c}{\texttt{result}}           \\ \cmidrule(lr){2-3} \cmidrule(lr){4-5} \cmidrule(lr){6-7} \cmidrule(lr){8-9}
& MAP   & NDCG$_{\%20}$     &  MAP  & NDCG$_{\%20}$ & MAP   & NDCG$_{\%20}$     &  MAP  & NDCG$_{\%20}$  \\\toprule
\textsc{MPNet-1B}            & 34.64 & 54.94 & 41.06 & 65.86 & 27.21 & 42.48 & 36.07 & 54.94\\
\textsc{SentBert-PP}            & 26.77 & 48.57 & 35.43 & 60.80 & 16.19 & 33.40 & 29.16 & 48.57\\
\textsc{SentBert-NLI}           & 25.23 & 45.39 & 30.78 & 54.23 & 15.02 & 31.10 & 30.27 & 45.39\\
\textsc{UnSimCSE-BERT}          & 24.45 & 42.59 & 30.03 & 51.59 & 14.82 & 31.23 & 28.76 & 42.59\\
\textsc{SuSimCSE-BERT}          & 23.24 & 43.45 & 30.52 & 55.22 & 13.99 & 30.88 & 25.58 & 43.45\\
\texttt{CoSentBert}             & 28.95 & 50.68 & 35.78 & 61.27 & 19.27 & 38.77 & 32.15 & 50.68\\
\textsc{ICTSentBert}            & 28.61 & 48.13 & 35.93 & 59.80 & 15.62 & 35.91 & 34.42 & 48.13\\[2pt]\midrule
\textsc{otMPNet-1B}             & 36.41 & 56.91 & 43.23 & 67.60 & 28.69 & 43.49 & 37.76 & 60.30\\
\textsc{Specter}                & 34.23 & 53.28 & 43.95 & 66.70 & 22.44 & 37.41 & 36.79 & 56.67 \\
\textsc{SciNCL}                & 39.37 & 59.24 & 49.64 & 70.02 & 27.14 & 46.61 & 41.83 & 61.70\\
\textsc{Specter-CoCite$_\mathtt{Scib}$}   & $\underset{\pm 1.48}{37.90}$ & $\underset{\pm 1.9}{58.16}$ & $\underset{\pm 2.51}{48.40}$ & $\underset{\pm 2.71}{68.71}$ & $\underset{\pm 0.96}{26.95}$ & $\underset{\pm 0.74}{46.79}$ & $\underset{\pm 2.17}{38.93}$ & $\underset{\pm 3.58}{59.68}$\\
\textsc{Specter-CoCite$_\mathtt{Spec}$}   & $\underset{\pm 0.73}{37.39}$ & $\underset{\pm 0.86}{58.38}$& $\underset{\pm 1.2}{49.99}$ & $\underset{\pm 1.16}{70.03}$ & $\underset{\pm 0.53}{25.60}$ & $\underset{\pm 1.35}{45.99}$ & $\underset{\pm 0.86}{37.33}$ & $\underset{\pm 1.02}{59.95}$\\\midrule
\textsc{ts\aspect}$_\mathtt{Spec}$             & $\underset{\pm 0.93}{40.26}$ & $\underset{\pm 0.67}{60.71}$ & $\underset{\pm 1.59}{49.58}$ & $\underset{\pm 1.74}{70.22}$ & $\underset{\pm 1.71}{28.86}$ & $\underset{\pm 1.72}{48.20}$ & $\underset{\pm 0.54}{42.92}$ & $\underset{\pm 0.28}{64.39}$\\
\textsc{ot\aspect}$_\mathtt{Spec}$          & $\underset{\pm 0.53}{\bf 40.79}$ & $\underset{\pm 0.52}{\bf 61.41}$ & $\underset{\pm 1.52}{50.56}$ & $\underset{\pm 1.42}{71.04}$ & $\underset{\pm 0.92}{27.64}$ & $\underset{\pm 0.1}{46.46}$ & $\underset{\pm 1.57}{\bf 44.75}$ & $\underset{\pm 0.99}{\bf 67.38}$\\
\textsc{ts+ot\aspect}$_\mathtt{Spec}$     & $\underset{\pm 0.71}{40.26}$ & $\underset{\pm 0.58}{60.86}$ & $\underset{\pm 1.18}{\bf 51.79}$ & $\underset{\pm 1.28}{70.99}$ & $\underset{\pm 3.21}{26.68}$ & $\underset{\pm 2.45}{47.60}$ & $\underset{\pm 0.21}{43.06}$ & $\underset{\pm 0.19}{64.82}$\\
\bottomrule
\end{tabular}
\caption{Test set results for baseline and proposed methods on \textsc{CSFCube}, an expert annotated fine-grained similarity dataset of computer science papers. Our approaches outperform strong prior models  \textsc{ot}/\textsc{MPNet-1B} and  \textsc{Specter} by 5-6 points, and the concurrently introduced \textsc{SciNCL} model by 1.5-2 points aggregated across queries. Metrics
(MAP, NDCG$_{\%20}$) are computed based on a 2-fold cross-validation and averaged over three re-runs of models. Standard deviations are 
below run averages. Here, \textsc{ts\aspect}: Text supervised single-match method, \textsc{ot\aspect}: Optimal Transport multi-match method and \textsc{ts+ot\aspect}: Multi-task multi aspect method.}
\label{tab-model-results-compsci-full}
\end{table*}
\begin{table*}[t]
\centering
\footnotesize
\scalebox{0.85}{
\begin{tabular}{@{}lcccc@{}} \toprule
& \multicolumn{2}{c}{\textsc{treccovid}$_\mathtt{RF}$}   & \multicolumn{2}{c}{\textsc{relish}}        \\ \cmidrule(lr){2-3} \cmidrule(lr){4-5} 
& MAP   & NDCG$_{\%20}$    &  MAP  & NDCG$_{\%20}$  \\\midrule
\textsc{MPNet-1B}  & 17.35 & 43.87 & 52.92 & 69.69\\\toprule
\textsc{SentBert-PP}  & 11.12 & 34.85 & 50.80 & 67.35\\
\textsc{SentBert-NLI} & 13.43 & 40.78 & 47.02 & 63.56\\
\textsc{UnSimCSE-BERT}  & 9.85  & 34.27 & 45.79 & 62.02\\
\textsc{SuSimCSE-BERT}  & 11.50 & 37.17 & 47.29 & 63.93\\
\texttt{CoSentBert}     & 12.80 & 38.07 & 50.04 & 66.35\\
\textsc{ICTSentBert}    & 9.80  & 33.62 & 47.72 & 63.71\\\midrule
\textsc{otMPNet-1B}     & 27.46 & 58.70 & 57.46 & 74.64\\
\textsc{Specter}        & 28.24 & 59.28 & 60.62 & 77.20\\
\textsc{SciNCL}        & 28.73 & 59.16 & 62.09 & 78.72\\
\textsc{Specter-CoCite$_\mathtt{Scib}$}    & $\underset{\pm 0.87}{30.60}$ & $\underset{\pm 0.95}{62.07}$ & $\underset{\pm 0.32}{61.43}$ & $\underset{\pm 0.1}{78.01}$\\
\textsc{Specter-CoCite$_\mathtt{Spec}$}    & $\underset{\pm 0.25}{28.59}$ & $\underset{\pm 0.36}{60.07}$ & $\underset{\pm 0.24}{61.43}$ & $\underset{\pm 0.23}{77.96}$\\\midrule
\textsc{ts\aspect}$_\mathtt{Spec}$            & $\underset{\pm 0.45}{26.24}$ & $\underset{\pm 0.65}{56.55}$ & $\underset{\pm 0.51}{61.29}$ & $\underset{\pm 0.42}{77.89}$\\
\textsc{ot\aspect}$_\mathtt{Spec}$         & $\underset{\pm 0.53}{\bf 30.92}$ & $\underset{\pm 0.67}{\bf 62.23}$ & $\underset{\pm 0.29}{62.57}$ & $\underset{\pm 0.26}{78.95}$\\
\textsc{ts+ot\aspect}$_\mathtt{Spec}$    & $\underset{\pm 0.71}{30.90}$ & $\underset{\pm 0.7}{62.18}$ & $\underset{\pm 0.16}{\bf 62.71}$ & $\underset{\pm 0.15}{\bf 79.18}$\\ \bottomrule
\end{tabular}
}
\caption{Test set results for baseline and proposed methods on \textsc{treccovid}$_{\mathtt{RF}}$ and \textsc{relish}, expert annotated abstract similarity datasets of biomedical papers. Our approaches outperform or match a strong prior model, \textsc{Specter}, and the concurrently introduced \textsc{SciNCL} by 2-3 points across metrics (MAP, NDCG$_{\%20}$). These are computed as averages over three model re-runs. Standard deviations are below run averages. Method names map similarly to Table \ref{tab-model-results-compsci-full}.}
\label{tab-model-results-biomed-full}
\end{table*}
\textbf{A3.\ Can single-match models be learned without co-citation contexts?} While our model for single matches leverages weak textual supervision from co-citation contexts, we ask if these models can be learned in the absence of this supervision. We answer this by training a simpler model, \textsc{max\aspect}, which finds the maximally aligned aspects between documents using the representations from $\mathtt{BERT}_{\theta}$ alone, giving us $f_{\mathtt{Max}}(p,p')= \texttt{max}_{i, j}\mathbf{D}$. To examine the role of $\mathtt{BERT}_{\theta}$ we compare performance with different initializations, with \textsc{Specter} presenting a initial model fine-tuned for similarity vs \textsc{SciBert} which isnt fine-tuned for text similarity.

We note the following from the results in Table \ref{tab-scib-spec-finegrained}: \textsc{max\aspect} sees a dependence on the underlying encoder, a \textsc{SciBert} initialization nearly always sees poorer performance -- only seeing performance competitive with \textsc{ts\aspect} when initialized with \textsc{specter}. This is reasonable given that this model must bootstrap fine-grained similarity while only relying on the encoder induced similarity. In cases where \textsc{max\aspect} matches performance of \textsc{ts\aspect} it sees larger performance differences across runs which may also be explained by the dependence on the initialization. Finally, \textsc{ts\aspect} consistently sees similar or better performance with varying initialization, indicating the value of our text supervised method.

\section{Extended Results}
Tables \ref{tab-model-results-compsci}, \ref{tab-model-results-biomed} in \S\ref{sec-main-results} omit presentation of standard deviations across runs for the proposed approaches for brevity. We include these in Tables \ref{tab-model-results-compsci-full} and \ref{tab-model-results-biomed-full}.

\section{Extended Related Work}
\label{appendix-relwork}
A range of modeling approaches in multi-instance learning, models leveraging textual supervision, and optimal transport resemble elements of our approach. We describe these next.

\textit{Multi-instance Learning:} Our work applies MIL for learning fine-grained similarity, while prior work has most often been applied to classification or regression tasks \cite{hope2016ballpark,hope2018ballpark,ilse2018attmi, angelidis2018multiple}. Our work bears resemblance to an application of MIL in content based image retrieval \cite{Song2019PVSE}, where MIL is applied to learn alignments between image and text aspects.

\textit{Textual Supervision:} Our use of co-citation text as a source of textual supervision draws on other work leveraging textual justifications of labels as a source of supervision for classification tasks \cite{hancock2018training, murty2020expbert, Hanjie2022semsup}, with recent concurrent work of Hanjie et al.\ including an overview of this line of work. Our co-citation contexts may be considered justifications for similarity of co-cited papers. \citet{Allen2020LitGen} presents work in a biomedical literature recommendation task, where human justifications of a relevance label are used to identify unigram features indicative of the label and train a recommendation model.

\textit{Optimal Transport:} Our use of optimal transport draws on other recent work in learning sparse alignments between texts \cite{swanson2020rationalizing, tam2019optimal}. Work of \citet{swanson2020rationalizing} learns sparse \emph{binary} alignments for sentence and document similarity tasks to rationalize decisions and \citet{tam2019optimal} leverage sparse soft alignments between characters for string similarity. \citet{kusner15docwmd} uses alignment based on word embeddings for document classification tasks using a K-nearest neighbors method. However, applying OT at the word level in scientific documents would lead to a large increase in computational complexity.

\end{document}